\documentclass[conference, a4paper]{IEEEtran}
\IEEEoverridecommandlockouts
\usepackage{cite}
\usepackage{amsmath,amssymb,amsfonts}
\usepackage{algorithmic}
\usepackage{graphicx}
\usepackage{textcomp}
\usepackage{booktabs}
\usepackage{adjustbox}
\usepackage{multirow}
\usepackage{float}
\usepackage{subfigure}
\usepackage{placeins}
\usepackage{array}
\usepackage{pifont}
\usepackage{caption}
\usepackage{xcolor}
\begin{document}

\title{Cross-modal Context Fusion and Adaptive Graph Convolutional Network for Multimodal Conversational Emotion Recognition}

\author{
\IEEEauthorblockN{1\textsuperscript{st} Junwei Feng}
\IEEEauthorblockA{\textit{School of Astronautics} \\
\textit{Northwestern Polytechnical University}\\
Xi'an, China \\
fjw@mail.nwpu.edu.cn}
\and
\IEEEauthorblockN{2\textsuperscript{nd} Xueyan Fan}
\IEEEauthorblockA{\textit{School of Software} \\
\textit{Northwestern Polytechnical University}\\
Xi'an, China \\
fanxueyan@mail.nwpu.edu.cn}
}


\maketitle

\begin{abstract}
Emotion recognition has a wide range of applications in human-computer interaction, marketing, healthcare, and other fields. In recent years, the development of deep learning technology has provided new methods for emotion recognition. Prior to this, many emotion recognition methods have been proposed, including multimodal emotion recognition methods, but these methods ignore the mutual interference between different input modalities and pay little attention to the directional dialogue between speakers. Therefore, this article proposes a new multimodal emotion recognition method, including a cross modal context fusion module, an adaptive graph convolutional encoding module, and an emotion classification module. The cross modal context module includes a cross modal alignment module and a context fusion module, which are used to reduce the noise introduced by mutual interference between different input modalities. The adaptive graph convolution module constructs a dialogue relationship graph for extracting dependencies and self dependencies between speakers. Our model has surpassed some state-of-the-art methods on publicly available benchmark datasets and achieved high recognition accuracy.
\end{abstract}


\section{Introduction}

Emotion Recognition in Conversation (ERC) \cite{ERC1,ERC2,ERC3,ERC4}, as a significant research area in artificial intelligence, holds immense application potential in fields such as human-computer interaction \cite{human-computer}, marketing \cite{marketing}, and healthcare \cite{Healthcare}. With the rapid advancement of deep learning technologies, ERC methods have witnessed remarkable innovation and progress \cite{survey}. Among these, multimodal emotion recognition approaches \cite{MERC}, which integrate information from multiple modalities, have gained significant attention due to their ability to comprehensively and accurately capture emotional expressions. In everyday interactions, emotional expressions are often conveyed through a combination of modalities, including language, facial expressions, and vocal tone. These modalities are inherently complementary and interdependent, offering rich emotional context when combined.

Despite their promise, existing multimodal emotion recognition methods \cite{weng2024enhancing, li2024lr, shen2023triplet} face notable challenges. First, integrating information from multiple modalities often introduces noise due to mutual interference, which can negatively impact recognition accuracy. Second, in conversational scenarios, these methods frequently overlook the bidirectional dependencies and intricate relationships between speakers, which are crucial for understanding emotional dynamics. Consequently, the inability to fully explore speaker relationships and dialogue context limits the depth and effectiveness of current models in capturing emotional interactions. 
Recently, diffusion models \cite{shen2024imagdressing, shen2024imagpose, shen2024boosting} have shown promise in mitigating such challenges by leveraging progressive noise reduction to refine features across multiple modalities. Their ability to model complex dependencies and generate context-aware representations offers potential advantages for capturing intricate emotional dynamics in multimodal and conversational settings.

To address these challenges, this paper proposes a novel multimodal emotion recognition framework that leverages cross-modal context fusion and adaptive graph convolutional networks to enhance performance. The proposed method consists of three key components: a cross-modal context fusion module, an adaptive graph convolutional encoding module, and an emotion classification module. The cross-modal context fusion module reduces noise by aligning and integrating contextual information across modalities, while the adaptive graph convolutional encoding module constructs a dialogue relationship graph to capture speaker dependencies and conversational directionality. Finally, the emotion classification module decodes these enriched features to classify emotions. Experimental results on publicly available ERC datasets demonstrate that the proposed model outperforms state-of-the-art methods, offering a new perspective for advancing multimodal emotion recognition research.
Our main contributions are summarized as follows:

\begin{itemize}
    \item We propose a novel multimodal emotion recognition framework that achieves state-of-the-art performance on two widely used ERC benchmark datasets.
    \item We design a cross-modal alignment module to reduce noise caused by mutual interference between different input modalities, improving the effectiveness of multimodal fusion.
    \item We introduce a multi-task learning-based loss function that enables the model to simultaneously handle coarse-grained and fine-grained emotion recognition tasks, enhancing its overall performance.
\end{itemize}

\begin{figure*}[t]
    \centering
    \includegraphics[width=1.0\textwidth]{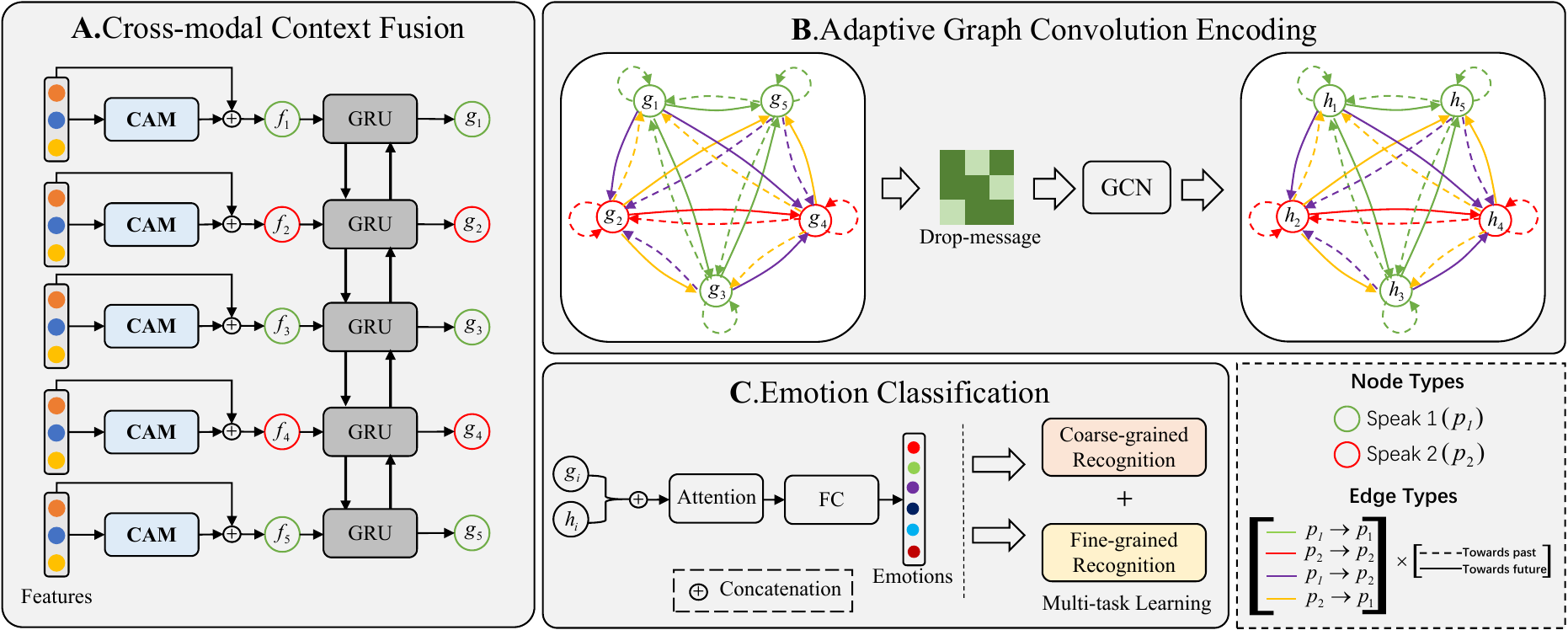} 
    \caption{Overall framework of our proposed method.The model consists of three key steps: \textbf{A. Cross-Modal Context Fusion}. Initially, the extracted features are processed through the Cross-Modal Alignment Module (CAM) to obtain enhanced information between modalities. After fusion, the features are further integrated with a bidirectional GRU to achieve deeper contextual feature fusion. \textbf{B. Adaptive Graph Convolutional Encoding}. In this step, speakers are modeled as a graph structure based on their conversational relationships. By processing through drop-message and graph convolutional network, the model effectively extracts dependencies among speakers and the directionality of the conversation. \textbf{C. Emotion Classification}. The encoded features are decoded and mapped to the dimensions of classification labels in this step. The model employs a multi-task learning training paradigm, with the loss function being the sum of losses for coarse-grained and fine-grained emotion classification.}
    \label{fig:MERC-GCN}
\end{figure*}

\section{Related work}
\label{sec:related work}

\subsection{Emotion Recognition in Conversation}

With the widespread use of social media and smart devices, a vast amount of data is generated in daily life, including text, images, and audio. These data contain rich emotional information, such as emotional states, reactions, and expressions. Consequently, Emotion Recognition in Conversations (ERC) has become an important research area. ERC can be applied not only in natural language processing, computer vision, and speech recognition but also provides effective solutions for human-computer interaction, sentiment analysis, and public opinion monitoring. With the advancement of deep learning technologies, numerous ERC methods based on deep learning have emerged. A model based on LSTM\cite{bclstm} was proposed to capture contextual information from the surrounding environment within the same video, aiding the classification process. The CMN\cite{CMN} conversational memory network was introduced, leveraging contextual information from conversational history. This framework employs a multimodal approach, including audio, visual, and textual features, with gated recurrent units to model each speaker's past utterances as memories. These memories are then merged through attention-based jumps to capture dependencies between speakers. A DialogueRNN\cite{DialogueRNN} model based on recurrent neural networks was developed to track the states of various parties throughout the conversation and use this information for emotion classification. The DialogueGCN\cite{DialogueGCN}, a graph convolutional neural network-based ERC method, was first proposed, focusing solely on textual features. A new model, MMGCN\cite{MMGCN}, based on multimodal fusion graph convolutional networks, was introduced, which can effectively utilize multimodal dependencies and model dependencies between and within speakers. However, this direct fusion approach may lead to redundant information and loss of heterogeneous information. 

\subsection{Graph Neural Network}

Convolutional neural networks (CNNs) have been widely used for extracting image features \cite{shen2023triplet}. However, CNNs exhibit inherent limitations when handling graph-structured data, as they are primarily designed for Euclidean space data. To address these challenges, graph neural networks (GNNs) \cite{shen2023git, shen2023pbsl} have emerged as a powerful alternative, enabling effective learning and inference in non-Euclidean domains. Unlike traditional deep learning models, which focus on processing vectors and matrices, GNNs leverage the topological structure of graphs and the relationships between nodes to capture complex dependencies.

Several GNN architectures have been proposed, including GCN \cite{GCN}, GraphSAGE \cite{GraphSAGE}, and GAT \cite{GAT}, each offering unique approaches to graph-based learning. The core idea of GCN is to generalize convolution operations from Euclidean space to graph structures. In traditional CNNs, convolutional operations extract local features via sliding windows. In contrast, GCN performs feature aggregation by combining information from neighboring nodes in the graph. Specifically, GCN updates each node's representation by applying a linear combination of its features and those of its neighbors, weighted by a learnable matrix. GCN's strengths include parameter sharing, adaptive aggregation, node embedding representation, and enhanced predictive capabilities.
GAT introduces an attention mechanism to GNNs, assigning different weights to neighbor nodes during feature aggregation. This mechanism dynamically adjusts the importance of neighbors based on their connections, enabling GAT to better adapt to diverse graph structures and capture a broader range of information. This flexibility makes GAT particularly effective in scenarios where certain nodes contribute more significantly to the task at hand.
GraphSAGE, on the other hand, adopts a sampling-based approach to efficiently learn node representations. Instead of aggregating all neighbor nodes like GCN, GraphSAGE samples a fixed number of neighbors and aggregates their features to approximate the global structure of the graph. This approach significantly reduces computational complexity, making it suitable for large-scale graph data. By allowing different sampling strategies and aggregation methods, GraphSAGE can adapt to various graph structures and capture richer contextual information.

\section{Method}
\label{sec:method}

The proposed model,named MERC-GCN,is designed for multimodal emotion recognition in conversations. The model consists of three steps: cross-modal context fusion, adaptive graph convolutional encoding, and emotion classification. The overall framework is illustrated in Fig.\ref{fig:MERC-GCN}.

\subsection{Problem Definition}

Assume there are \( M \) speakers in a conversation, with the sequence of utterances represented as \( u_1, u_2, \cdots, u_N \), where each utterance \( u_i \) is spoken by speaker \( p_s(u_i) \). Each utterance contains three emotional modalities \( u_i^V, u_i^A, u_i^T \), where \( V \), \( A \), and \( T \) represent information from visual, audio, and textual sources, respectively. Our task is to predict the emotional category \( y_i \) of the speaker corresponding to each utterance \( u_i \).

\subsection{Preprocessing: Unimodal Feature Extraction}

\textbf{Text Modality}: RoBERTa\cite{RoBERTa} is a variant of BERT\cite{BERT} that employs more efficient pre-training methods, making it a more robust pre-trained language model than BERT. In this paper, RoBERTa is used to encode text information into a 200-dimensional feature vector. All text features are denoted as \( U^T \).

\textbf{Audio Modality}: openSMILE\cite{Opensmile} (open-source Speech and Music Interpretation by Large-space Extraction) is an open-source toolkit for audio feature extraction and classification of speech and music signals. openSMILE is widely used in affective computing for automatic emotion recognition. openSMILE performs the following four types of feature extraction operations: signal processing, data processing, audio features (low-level), and functionals. In this paper, openSMILE is used to encode audio information into a 100-dimensional feature vector. All audio features are denoted as \( U^A \).

\textbf{Visual Modality}: DenseNet\cite{DenseNet} is a type of CNN network whose basic concept is similar to ResNet\cite{ResNet} but establishes dense connections between all preceding layers and subsequent layers, enabling feature reuse through connections across channels. CNN networks are better suited for capturing image features, and in this paper, DenseNet is used to encode video information into a 100-dimensional feature vector. All video features are denoted as \( U^V \).

\subsection{Method}
\subsubsection{Cross-modal Context Fusion}

Different modalities at the same time have correlations. If these are directly concatenated as input features to the network, the network might confuse the correlations between different modal features. Therefore, this paper uses a co-attention transformer (CT) \cite{CT}for cross-modal enhancement to learn distinct cross-modal correlated features.

\begin{figure}[t]
\centering
\subfigure[Cross-Modal Alignment Module.]{
    \includegraphics[width=0.5\linewidth]{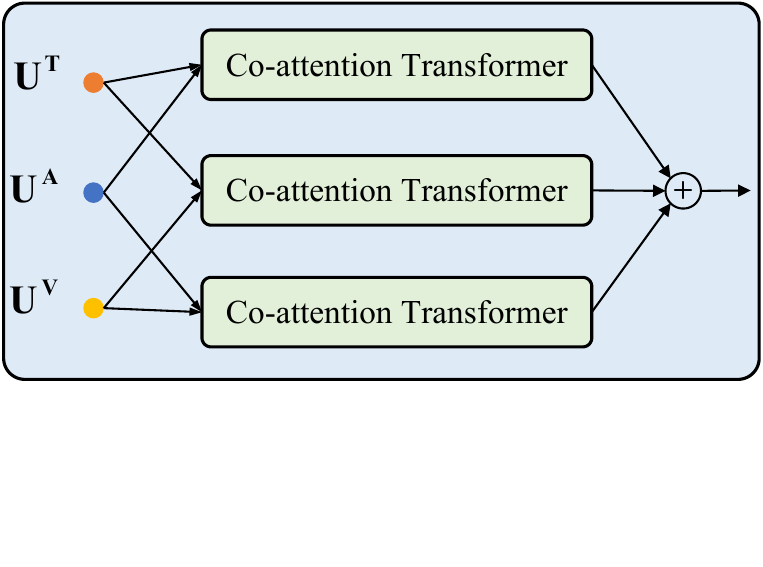}
}
\subfigure[Co-Attention Transformer.]{
    \includegraphics[width=0.4\linewidth]{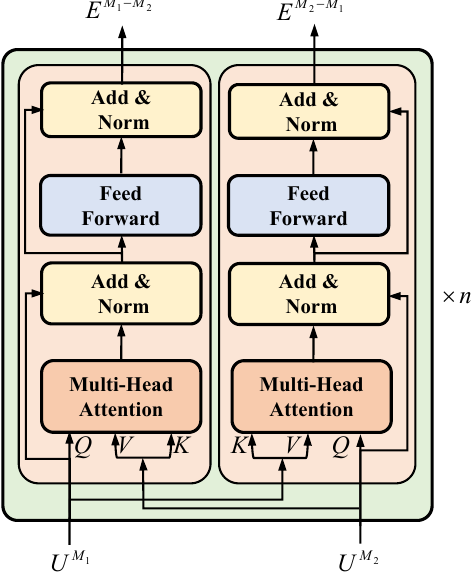}
}
\caption{(a) \textbf{Cross-Modal Alignment Module (CAM).} The input modalities are processed pairwise through the co-attention mechanism module, learning enhanced cross-modal representations and performing fusion. (b) \textbf{Co-Attention Transformer.}This module enhances the model's ability to capture inter-modal dependencies, leading to more accurate and context-aware representations.}
\label{fig:CAM}
\vspace{-3mm}
\end{figure}

As shown in the Fig.\ref{fig:CAM}, each CT learns cross-modal representations between two modalities; thus, three co-attention transformers are required to learn cross-modal representations for each pair of the three modalities in the ERC task. Each CT block consists of two identical parts, left and right, with symmetrical input. In the left part, one input modality is used as the query, while the other modality is used as the key and value, with the latter weighted and summed under the guidance of the former. The right part of the CT block undergoes a symmetrical process simultaneously. This entire process repeats \( T \) times, outputting the mutual cross-modal representations of the two input modalities.

Co-attention transformer reduces the semantic gap between modalities and enhances shared features between them, achieving modality alignment and reducing noise in the input modalities. The entire process is mathematically represented as:

\begin{equation}
\text{MultiHead}(Q, K, V) = (\text{head}_1 \oplus \dots \oplus \text{head}_h) W^O,
\end{equation}
\begin{equation}
Q_i = QW_i^Q, \quad K_i = KW_i^K, \quad V_i = VW_i^V,
\end{equation}
\begin{equation}
\text{head}_i = \text{Att}(Q_i, K_i, V_i) = \text{softmax}\left(\frac{Q_i K_i^T}{\sqrt{d_h}}\right) V_i.
\end{equation}

Here, \( \oplus \) denotes the concatenation operation. \( Q, K, V \in \mathbb{R}^{L \times d_{model}} \) represent two of the input modalities \( U^A,U^V, U^T \), as previously described. \( L \) is the length of the input feature vector of the corresponding modality. \( W^O \in \mathbb{R}^{hd_h \times d_{model}}, W_i^Q, W_i^K, W_i^V \in \mathbb{R}^{d_{model} \times d_h} \) are learnable hyperparameters. \( d_{model} \) and \( h \) are inherent hyperparameters of the model, and in this paper, \( h = 8 \) and \( d_h = d_{model}/h = 64 \).

The feedforward neural network consists of two linear layers, mathematically represented as:

\begin{equation}
\text{FFN}(X) = \sigma(\sigma(X W_1 + b_1) W_2 + b_2).
\end{equation}

Where \( X \in \mathbb{R}^{L \times d_{model}} \) is the output after the first residual connection and layer normalization in the CT block, and \( \sigma \) represents the activation function, with ReLU being used in this paper. The CT block is stacked \( T \) times, with the output of the previous CT block serving as the input to the next, achieving enhanced representation. This entire step can be described mathematically as follows:

\begin{equation}
E^{T-A}, E^{A-T} = CT(U^{T}, U^{A}),
\end{equation}
\begin{equation}
E^{V-A}, E^{A-V} = CT(U^{A}, U^{V}),
\end{equation}
\begin{equation}
E^{T-V}, E^{V-T} = CT(U^{T}, U^{V}).
\end{equation}

CT represents the co-attention transformer, constructed by \( T \) stacked co-attention transformer blocks. \( E^{T-A} \) denotes the cross-modal representation of the text modality relative to the visual modality, and so on.

We concatenate the learned cross-modal correlated features with the original features to prepare for the next step of context feature fusion. This is mathematically represented as:

\begin{equation}
\begin{aligned}
F = [&E^{T-A}, E^{A-T}, E^{V-A}, E^{A-V}, \\
     &E^{T-V}, E^{V-T}, U^{T}, U^{A}, U^{V}].
\end{aligned}
\end{equation}

For the \( i \)-th utterance, the features it carries are denoted as \( f_i \), so:

\begin{equation}
F = [f_1, f_2, \dots, f_N].
\end{equation}

Conversations occur sequentially, with contextual information flowing along this sequence. Based on this characteristic, we constructed a bidirectional gated recurrent unit (BiGRU) \cite{GRU} to capture contextual information. The input modality features include both the original modality features and the cross-modal correlated features, achieving fusion and interaction within the flow of contextual information. The specific mathematical formula is as follows:

\begin{equation}
g_i = \left[ \overrightarrow{\text{GRU}}(f_i, g_{i-1}), \overleftarrow{\text{GRU}}(f_i, g_{i+1}) \right].
\end{equation}

Here, \( g_i \) represents the feature after sequential context fusion. This step integrates sequential contextual modality features but does not yet account for speaker identity and inter-speaker dependencies. These aspects will be considered in the next step.

\subsubsection{Adaptive Graph Convolution Encoding}

We constructed a graph convolutional neural network to encode the relationships between speakers, thereby capturing both inter-speaker dependencies and self-dependencies.

First, we define the following symbols: based on a scenario with \( N \) utterances, we construct a directed graph \( \mathcal{G} = (V, \mathcal{E}, R, W) \), where nodes \( v_i \in V \) and \( r_{ij} \in R \) represent a directed edge from node \( v_i \) to node \( v_j \), and \( \alpha_{ij} \in W \) represents the weight of the directed edge \( r_{ij} \), with \( 0 \leq \alpha_{ij} \leq 1 \), \( i, j \in [1, 2, \cdots, N] \).

\begin{flushleft}
\textbf{1. Dialogue Graph Construction}
\end{flushleft}

\textbf{Nodes}: Each utterance \( u_i \) in the conversation represents a node \( v_i \in V \) in the graph. For each node \( i \in [1, 2, \cdots, N] \), we initialize it with the encoded sequential context feature vector \( g_i \). This vector serves as the feature of the node. After speaker-level encoding within the model, the sequential context feature vector is transformed into the corresponding speaker-level feature vector.

\textbf{Edges}: The construction of edges models the conversational relationships between speakers. Assuming each utterance is a vertex, it affects and is affected by all other vertices (including itself) to varying degrees. This relationship is represented by directed edges in a directed graph, where the influence of \( u_i \) on itself is represented by a directed edge from \( u_i \) to \( u_i \). This reflects, in practical terms, the inertia of the speaker themselves. However, using all \( N \) utterances to construct this directed graph results in a computational complexity of \( O(N^2) \), which can be very costly when there are many utterances. In practice, instead of using all utterances, we can consider only those within a certain time frame, representing a past context window of \( p \) utterances observed in the past and a future context window of \( f \) utterances to be observed in the future, constructing a directed graph with \( p+f \) vertices. Each vertex \( u_i \) has edges directed to the past \( p \) vertices and the future \( f \) vertices, representing its influence on past and future utterances. In this paper’s experiments, we set both the past and future context window sizes to 10, meaning the directed graph is constructed using 10 past and 10 future utterances.

\textbf{Edge Weights}: The weights of the edges are calculated using a similarity-based attention mechanism. The calculation method of the attention function ensures that for each vertex, the total weight of incoming edges sums to 1. Considering the past context window size \( p \) and the future context window size \( f \), the weight calculation is as follows:

\begin{equation}
\begin{aligned}
&\alpha_{ij} = \text{softmax} \left( g_i^T W_e [g_{i-p}, \dots, g_{i+f}] \right), \\
&\text{for } j = i - p, \dots, i + f.
\end{aligned}
\end{equation}

This ensures that the total weight contribution of the incoming edges for vertex \( v_i \) from vertices \( v_{i-p}, \cdots, v_{i+f} \) sums to 1. Different weight values represent the varying influence of the corresponding vertices.

\begin{flushleft}
\textbf{2. Graph Representation Learning}
\end{flushleft}
Before this step, the feature \( g \) is a multimodal fusion feature independent of speaker relationships, including text semantics and real-time representations of audio and video. Next, we use a graph convolutional network to perform a two-step feature transformation to extract representations of connections between speakers.

In the first step, we use one layer of GCN to aggregate neighborhood information of vertices, thereby initially encoding the directional nature of conversations between speakers. In this step, we use the DropMessage method to enhance the aggregation capability of GCN. We generate a mask matrix of the same size as the message matrix based on a Bernoulli distribution, where each element in the message matrix is dropped to a certain extent as determined by the corresponding value in the mask matrix. After applying dropmessage, the mathematical formula for the node features is:

\begin{equation}
\tilde{g}_i = 
\begin{cases} 
g_{[M]}, & v_i \in V_M, \\ 
g_i, & v_i \notin V_M.
\end{cases}
\end{equation}

where \( V_M \) denotes the masked nodes, \( g[M] \) represents the feature vector of the masked nodes, and \( \tilde{g}[M] \) represents the updated node features.

The mathematical formula for masked edges is:

\begin{equation}
\tilde{e}_{ij} = 
\begin{cases} 
e_{ij}^{[M]}, & \alpha_i \in E_M, \\ 
e_{ij}, & \alpha_i \notin E_M.
\end{cases}
\end{equation}

where \( \phi_M \) denotes the masked edges, \( e_{ij}[M] \) represents the weight of the masked edges, and \( \hat{e_{ij}[M]} \) denotes the updated edge weight.

The overall learning formula for this step is:

\begin{equation}
\begin{aligned}
h_i^{(1)} = &\sigma \left( \left( \sum_{k \in \mathcal{R}} \sum_{r \in R} \sum_{j \in N_i^r} \frac{\alpha_{ij}}{c_{i,r}} W_r^{(1)} \tilde{g}_j \right. \right. \\
& + \alpha_{ii} W_0^{(1)} \tilde{g}_i \Bigg) \cdot \tilde{e}_{ik} \Bigg).
\end{aligned}
\end{equation}

where \( \alpha_{ii} \) and \( \alpha_{ij} \) are the edge weights, and \( N_i^r \) is the neighborhood index of vertex \( i \) under relationship \( r \in R_c \). \( c_{i, r} \) is a normalization constant specific to the task and automatically learned in a gradient-based learning setup. \( \sigma \) is an activation function like ReLU, and \( W_r \) and \( W_0 \) are learnable transformation parameters.

In the second step, we apply GCN again to extract relationship features between vertices, reinforcing the extraction of features that capture the conversational relationships between speakers:

\begin{equation}
\begin{aligned}
h_i^{(2)} = &\sigma \left( \sum_{j \in N_i^r} W^{(2)} h_j^{(1)} + W_0^{(2)} h_i^{(1)} \right), \\
&\text{for } i = 1, 2, \dots, N.
\end{aligned}
\end{equation}

where \( W_c \) and \( W_0 \) are learnable parameters, and \( \sigma \) is an activation function.

This step constructs a graph model of conversational relationships between speakers, building upon the previous step's cross-modal context feature fusion to capture the conversational relationship features between speakers.

\subsubsection{Emotion Classification}
We fuse the context encoding vector with the speaker encoding vector and use an attention mechanism to learn the importance of different features:

\begin{equation}
h_i = [g_i, h_i^{(2)}],
\end{equation}

\begin{equation}
\beta_i = \text{softmax} \left( h_i^T W_{\beta} [h_1, h_2, \dots, h_N] \right),
\end{equation}

\begin{equation}
\bar{h}_i = \beta_i [h_1, h_2, \dots, h_N]^T.
\end{equation}

Finally, we feed the resulting features into an MLP for decoding. The softmax function outputs the final predicted probability distribution for each class, and we select the label corresponding to the highest probability as the prediction result:

\begin{equation}
l_i = \text{ReLU} \left( W_l \tilde{h}_i + b_l \right),
\end{equation}

\begin{equation}
P_i = \text{softmax} \left( W_{s_{\text{max}}} l_i + b_{s_{\text{max}}} \right),
\end{equation}

\begin{equation}
\hat{y}_i = \arg \max_k \left( P_i[k] \right).
\end{equation}

\subsection{Optimization Objective}

We use the categorical cross-entropy loss function as the objective function for training. We adopt a multi-task learning strategy, with the loss function consisting of two parts that reflect the model’s learned emotional biases at both fine-grained and coarse-grained levels. Emotions are divided into coarse-grained and fine-grained categories. Taking the IEMOCAP\cite{IEMOCAP} dataset as an example, the fine-grained emotional labels are happy, excited, neutral, sad, angry, and frustrated. Among them, happy and excited ones are considered positive, neutral ones are still neutral, and others are negative, resulting in coarse-grained emotional labels.

\begin{table*}[t]
\centering
\caption{Performance comparison on IEMOCAP and MELD datasets for different emotion recognition models.}
\begin{adjustbox}{width=\textwidth}
\begin{tabular}{c c c c c c c c |c}
\hline
\multirow{2}{*}{Model} & \multicolumn{7}{c}{IEMOCAP} & \multicolumn{1}{|c}{MELD} \\ \cline{2-9} 
                       & Happy & Sad & Neutral & Angry & Excited & Frustrated & Average(w) & Average(w) \\ \hline
bc-LSTM \cite{bclstm}               & 32.63 & 70.34 & 51.14 & 63.44 & 67.91 & 61.06 & 59.58 & 56.80 \\ 
CMN \cite{CMN}                    & 30.38 & 62.41 & 52.39 & 59.83 & 60.25 & 60.69 & 56.56 & - \\ 
ICON \cite{ICON}                  & 29.91 & 64.57 & 57.38 & 63.04 & 63.42 & 60.81 & 59.09 & - \\ 
DialogueRNN  \cite{DialogueRNN}          & 33.18 & 78.80 & 59.21 & 65.28 & 71.86 & 58.91 & 63.40 & 57.66 \\ 
DialogueGCN \cite{DialogueGCN}          & 47.10 & 80.88 & 58.71 & 66.08 & 70.97 & 61.21 & 65.54 & 56.36 \\ 
MMGCN \cite{MMGCN}                 & 45.45 & 77.53 & 61.99 & 66.67 & 72.04 & 64.12 & 65.56 & 57.82 \\ 
MERC-GCN (ours)               & \textbf{68.90} & 78.12 & \textbf{66.48} & 58.33 & \textbf{79.66} & 62.01 & \textbf{68.98} & \textbf{62.54} \\ \hline
\end{tabular}
\end{adjustbox}
\label{tab:performance_comparison}
\end{table*}


The coarse-grained emotion loss function is:

\begin{equation}
L_C = - \frac{1}{\sum\limits_{s=1}^{N} c(s)} \sum_{i=1}^{N} \sum_{j=1}^{c(i)} \log P_{i,j} [y_{j,i}^{C}].
\end{equation}

The fine-grained emotion loss function is:

\begin{equation}
L_F = - \frac{1}{\sum\limits_{s=1}^{N} c(s)} \sum_{i=1}^{N} \sum_{j=1}^{c(i)} \log P_{i,j} [y_{j,i}^{F}].
\end{equation}

Our final training objective is:

\begin{equation}
\begin{aligned}
L &= \alpha L_C + (1 - \alpha)L_F + \lambda \lVert \theta \rVert \\
  &= - \frac{1}{\sum\limits_{s=1}^{N} c(s)} \sum_{i=1}^{N} \sum_{j=1}^{c(i)} \left\{ \alpha \log P_{i,j} [y_{i,j}^{C}] \right. \\
  &\quad + \left. (1 - \alpha) \log P_{i,j} [y_{i,j}^{F}] \right\} + \lambda \lVert \theta \rVert.
\end{aligned}
\end{equation}

where \( N \) is the number of conversations, \( c(s) \) is the number of utterances in conversation \( s \), \( P_{i,j} \) is the probability distribution of the predicted emotion label for utterance \( j \) in conversation \( i \),$\alpha$ is the coarse-grained loss weight.

\section{Experimental Setup}
\label{sec:experiment}

\subsection{Datasets}

We evaluate our model on two benchmark datasets: IEMOCAP\cite{IEMOCAP} and MELD\cite{MELD}. These two datasets are designed for emotion recognition and contain three modalities: text, video, and audio.

\textbf{IEMOCAP} consists of 10 hours of multimodal conversations performed by 10 actors. Each emotional conversation is carried out between two actors to simulate emotional communication in real-life situations. The dataset includes five emotion labels: Happy, Anger, Sadness, Neutral, and Excitement.

\textbf{MELD} contains 1,430 dialogue segments from the TV show "Friends," with each segment consisting of multiple dialogue turns. The dataset includes seven emotion labels: Anger, Disgust, Fear, Joy, Sadness, Surprise, and Neutral.

\subsection{Hyperparameters}

The experiments were conducted on an RTX 4090 GPU, with a batch size set to 32 and a total of 60 training epochs. The Adam optimizer was used with a learning rate of 0.005.

\section{Experimental Results}

\subsection{Comparison}

We compared the performance of our proposed MERC-GCN framework with state-of-the-art MMGCN and other baseline methods as shown in the table \ref{tab:performance_comparison}. On the IEMOCAP dataset, MERC-GCN achieved a new state-of-the-art accuracy of 68.98\%, which is about 3\% better than MMGCN and DialogueGCN, and at least 10\% better than all other models, outperforming SOTA methods in three emotional dimensions. Similarly, on the MELD dataset, MERC-GCN achieved a weighted accuracy of 62.54\% across four emotional dimensions, outperforming other baseline models. The reason for this gap lies in the inherent differences of the models. MERC-GCN, DialogueGCN, and DialogueRNN all attempt to extract speaker-level features, while other models usually focus solely on context information. Extensive research has shown  that speaker-level features are crucial for emotion recognition tasks, which is why algorithms that focus on speaker-level information tend to outperform those that neglect it.

Regarding the performance differences between MERC-GCN, DialogueGCN, and DialogueRNN, DialogueRNN uses Gated Recurrent Units (GRU) to extract speaker-level information, while DialogueGCN uses graph convolutional networks to overcome the issue of long sequence information propagation caused by the limitations of the recurrent encoder in DialogueRNN. We speculate that speaker-level information is often hidden in the interactions of the text, speech, and video modalities. Other algorithms only extract speaker-level information through text, which may result in insufficient use of all three modalities. This happens in real-world scenarios where there are inconsistencies between text and video at the speaker level, such as when the meaning conveyed by the text contrasts with the body language reflected in the video. In contrast, MERC-GCN extracts sufficient speaker-level information across multiple modalities and conversation relationships through cross-modal attention, thus overcoming the issue of single-modality speaker-level extraction.

\begin{figure}[t]
\centering
\subfigure[IEMOCAP]{
    \includegraphics[width=0.45\linewidth]{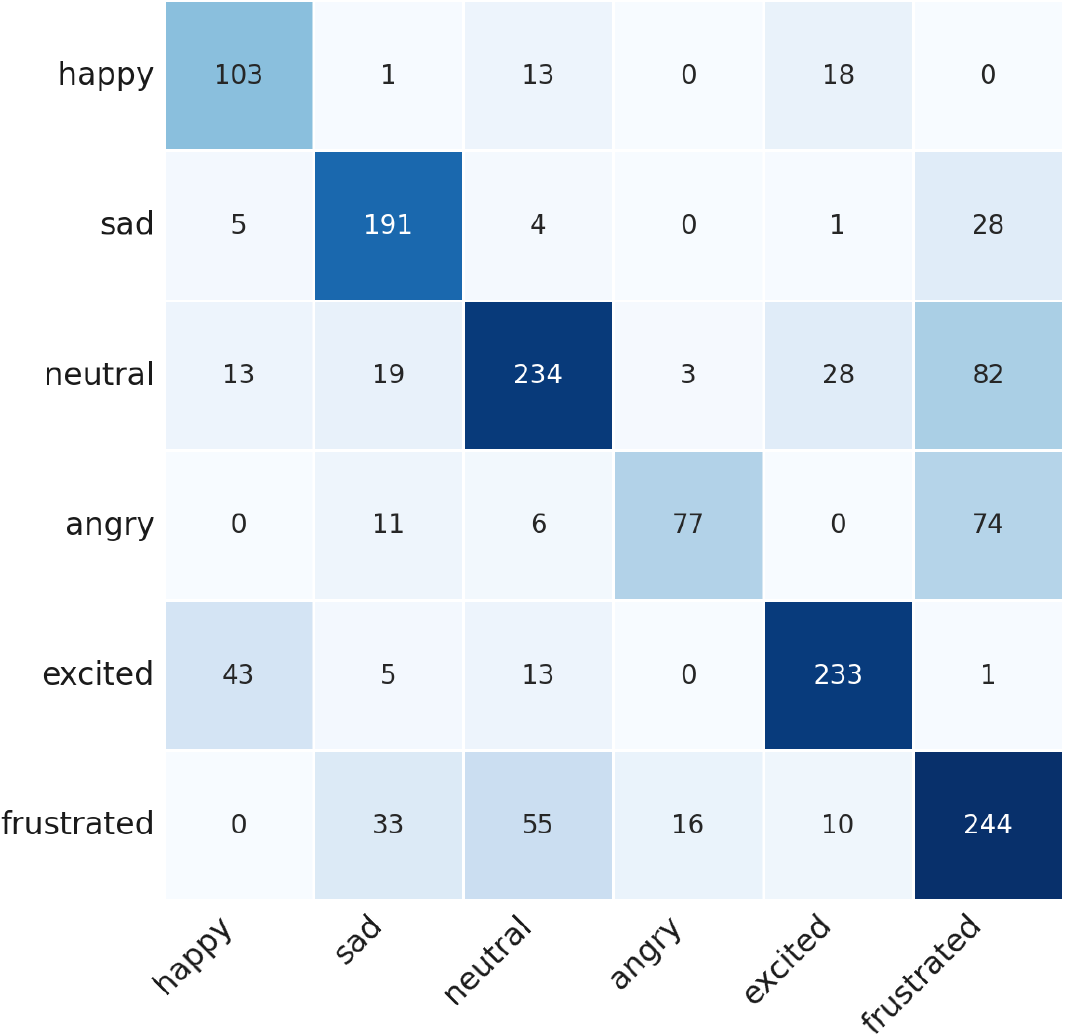}
}
\subfigure[MELD]{
    \includegraphics[width=0.45\linewidth]{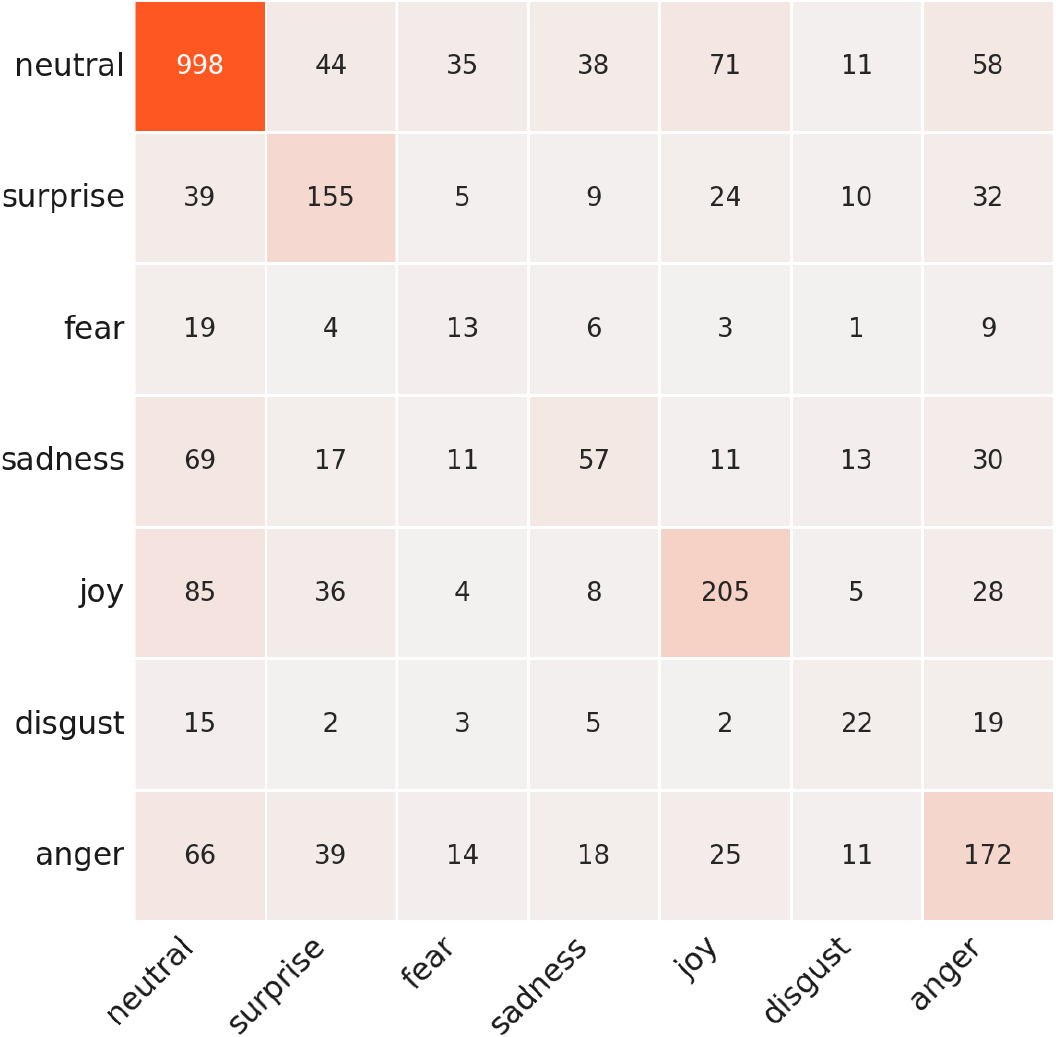}
}
\caption{Confusion matrix.In a confusion matrix, each row represents the actual class, and each column represents the predicted class. }
\label{fig:Confusion matrix}
\vspace{-3mm}
\end{figure}

Moreover, the standard deviations for DialogueGCN and DialogueRNN across different categories are 12.65 and 10.04, respectively, while our MERC-GCN has a standard deviation of only 7.83. This is due to the multi-task learning strategy, which merges categories or uses coarse-grained classification, making the model’s performance on each category more balanced during training.

Fig.\ref{fig:Confusion matrix} presents the confusion matrix of our model on two datasets. It can be seen that our model has a high recognition accuracy and is not easily confused on the same coarse-grained task, thanks to the training strategy we adopted for multi-task learning.

\subsection{Ablation Study and Analysis}

As shown in the table \ref{tab:Ablation Study},we conducted ablation experiments on different stages (i.e., cross-modal context fusion and adaptive graph convolutional encoder), as shown in the table. We found that the speaker-level encoder is slightly more important for overall performance. We speculate that relying solely on either cross-modal context fusion or the adaptive graph convolutional encoder may not fully capture the complexity of emotional expressions. The synergy of both components better models the emotions of different speakers, highlighting the importance of cross-modal context fusion and the adaptive graph convolutional encoder in dialogue emotion recognition.

\begin{table}[t]
\centering
\caption{Ablation Study.}
\begin{adjustbox}{width=0.8\linewidth}
\begin{tabular}{c c c c c}
\hline
Module A & Module B & F-score & Acc \\ \hline
\ding{55} & \ding{55} & 38.52 & 39.16 \\ 
\ding{55} & \checkmark & 66.25 & 67.31 \\ 
\checkmark & \ding{55} & 65.69 & 66.48 \\ 
\checkmark & \checkmark & \textbf{68.98} & \textbf{69.18} \\ \hline
\end{tabular}
\end{adjustbox}

\label{tab:Ablation Study}
\end{table}

\subsection{Hyperparameter Optimization}
\subsubsection{Context Fusion Encoding Model}

\begin{figure}[t]
\centering
\subfigure[IEMOCAP]{
    \includegraphics[width=0.46\linewidth]{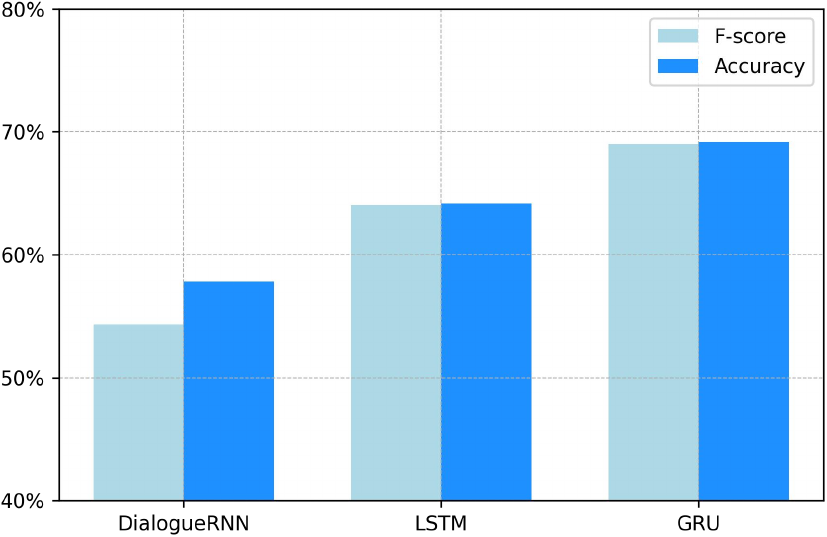}
}
\subfigure[MELD]{
    \includegraphics[width=0.46\linewidth]{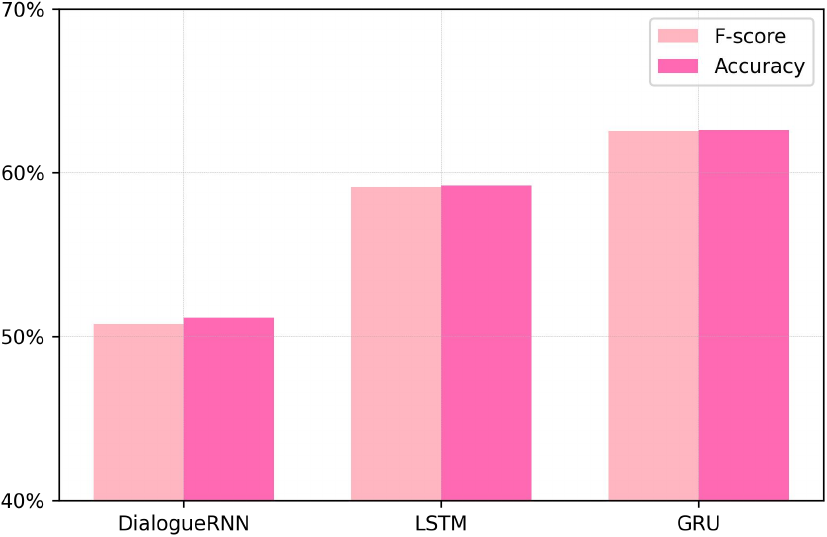}
}
\caption{Accuracy and F-score comparisons with different RNNs.The experiment indicates that among these RNNs, GRU performs the best on both the IEMOCAP and MELD datasets. Following GRU, LSTM yields the second-best results, while the DialogueRNN exhibits the poorest performance.}
\label{fig:RNNs}
\vspace{-3mm}
\end{figure}

We conducted ablation experiments on different context fusion models.As shown in Fig.\ref{fig:RNNs}, when the context fusion model used our GRU module, both the F-score and accuracy were better than those using DialogueRNN and LSTM, with the F-score being approximately 15\% higher than DialogueRNN and accuracy about 12\% higher. Compared to LSTM and DialogueRNN, the gated units used in GRU can more effectively capture contextual information. The update and reset gates in GRU better control the flow of information. Furthermore, GRU’s tolerance to noise and precise control of information flow make it perform more effectively in dialogue emotion recognition tasks.

\subsubsection{Multi-task Learning Hyperparameter Optimization}

\begin{figure}[t]
\centering
\subfigure[IEMOCAP]{
    \includegraphics[width=0.46\linewidth]{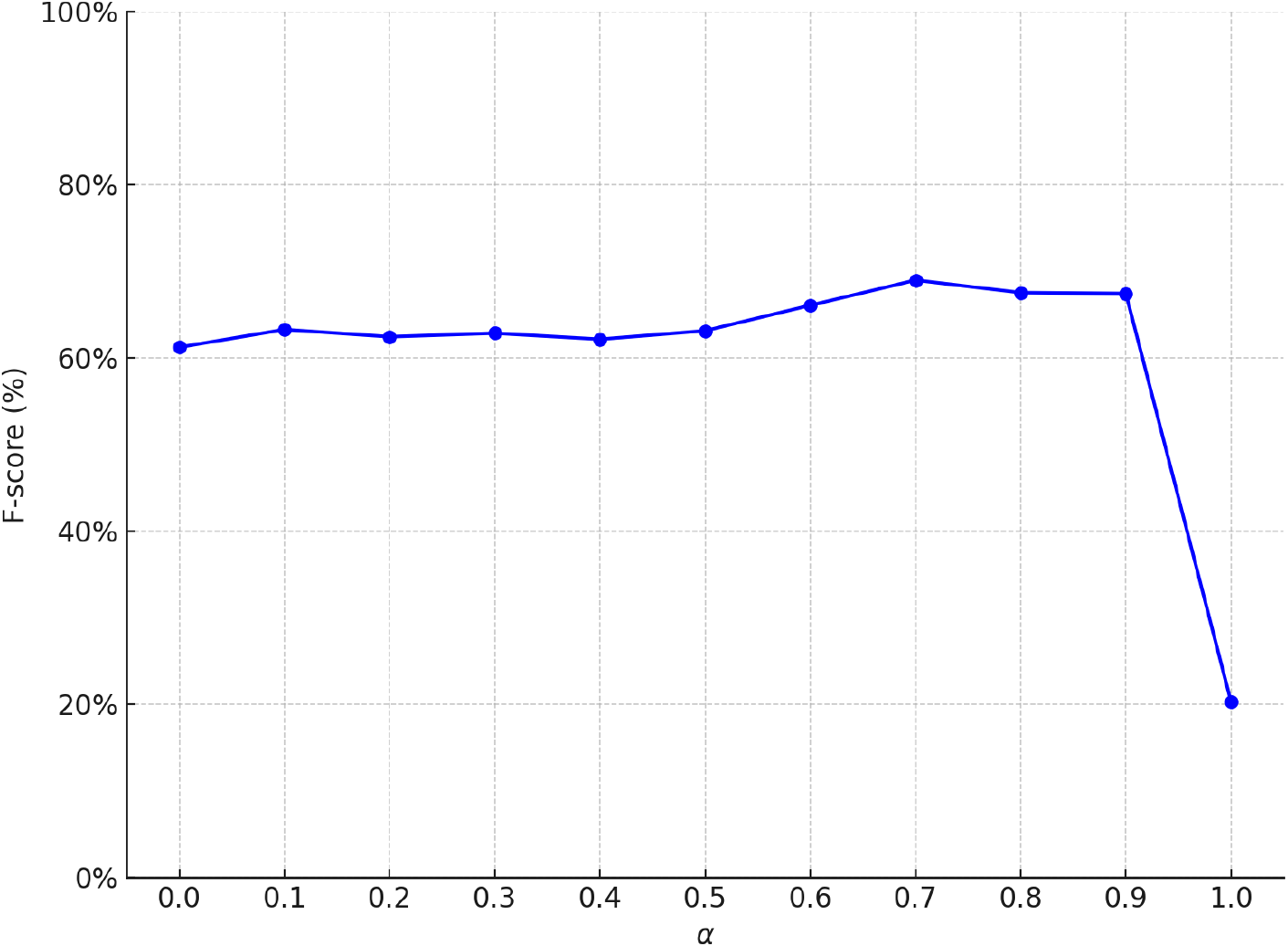}
}
\subfigure[MELD]{
    \includegraphics[width=0.46\linewidth]{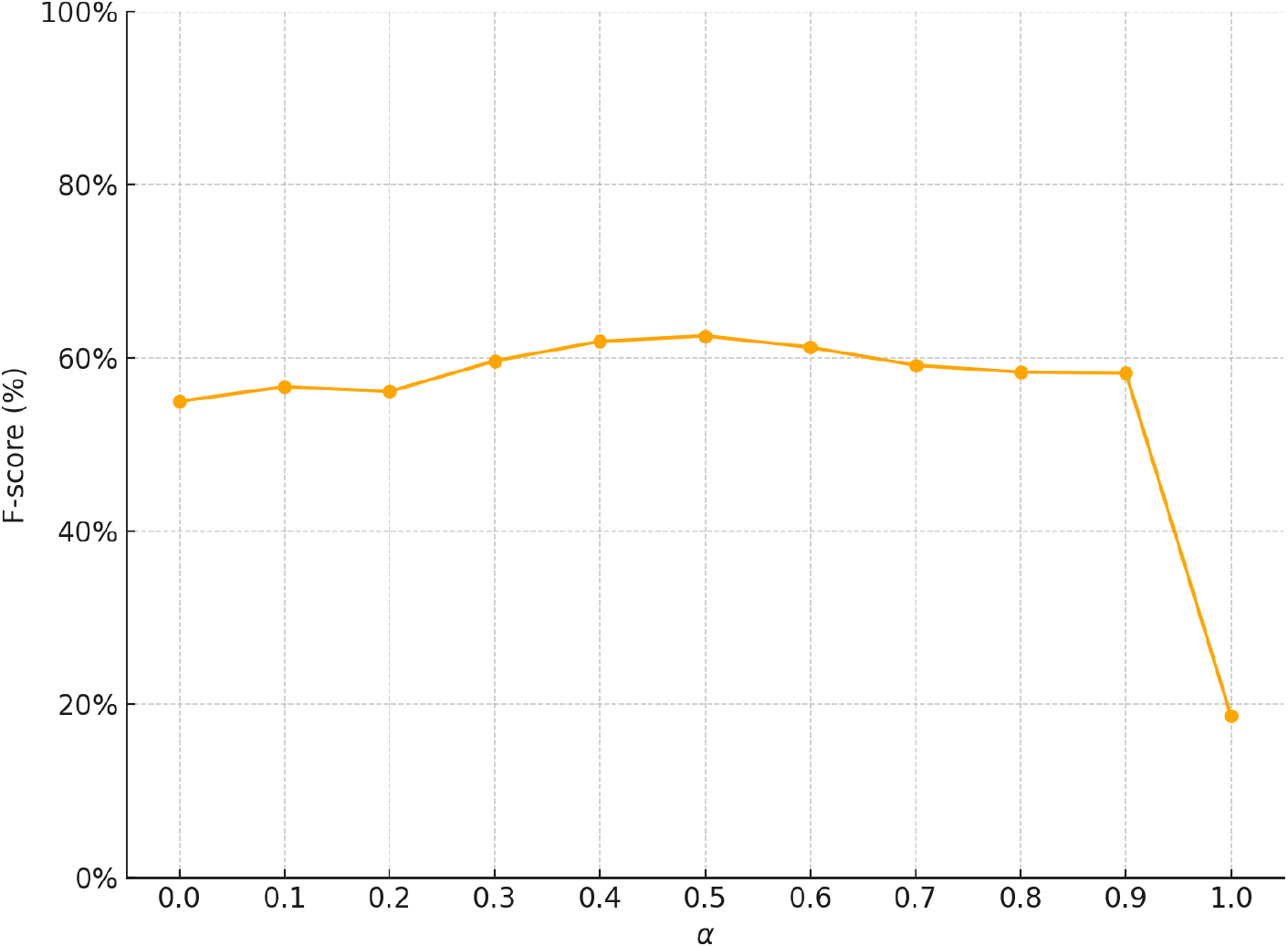}
}
\caption{Effect of parameter $\alpha$ on F-score.On the IEMOCAP dataset, the model has the highest F-score when the value of parameter $\alpha$ is 0.7, while on the MELD dataset, the model has the highest F-score when the value of parameter $\alpha$ is 0.5.}
\label{fig:alpha}
\vspace{-3mm}
\end{figure}

We conducted a comparison experiment on different coarse-grained weights with respect to the learning rate, as shown in the Fig.\ref{fig:alpha}. On the IEMOCAP dataset, when the coarse-grained weight was set to 0.7, both the F-score and accuracy were optimal, while on the MELD dataset, the optimal parameter was 0.5. This difference may be due to the class imbalance in the datasets. In IEMOCAP, the samples for the Anger, Happy, and Sadness labels are relatively abundant, while in MELD, there are more samples for Anger and Happy. When coarse-grained classification is not used at all, the model tends to predict the larger classes in the training set, thereby lowering overall accuracy. Merging classes or applying coarse-grained classification helps to reduce the imbalance between categories, making the model's performance on each category more balanced during training. The model performs best when the dataset distribution is imbalanced, as it helps the model fit the true labels more accurately when updating weights.

\subsubsection{Modality Ablation Experiment}

We conducted ablation experiments on different modalities of information, including individual modalities and pairs of combined modalities, as shown in the table \ref{tab:Modality}. The contribution of each modality to performance improvement varies, with the video modality making the greatest contribution, followed by audio, while the text modality has the least impact. For pairs of modalities, although theoretically they can achieve information complementarity, due to issues like information loss and modality alignment, the combination did not significantly improve performance and may have even caused interference. The model achieved the best performance when all three modalities were used together.

\begin{table}[t]
\centering
\setlength{\tabcolsep}{15pt} 
\caption{Performance metrics for different modality combinations}
\begin{tabular}{ccc}
\toprule
\textbf{Modality} & \textbf{F-score} & \textbf{Acc} \\
\midrule
T       & 65.31 & 65.41 \\
V       & 67.31 & 67.33 \\
A       & 66.30 & 66.50 \\
T-V     & 65.32 & 65.35 \\
T-A     & 65.87 & 66.24 \\
A-V     & 65.66 & 65.92 \\
T-A-V   & \textbf{68.98} & \textbf{69.18} \\
\bottomrule
\end{tabular}
\label{tab:Modality}
\end{table}

\section{Conclusion}
\label{sec:conclusion}

In this paper, we proposed cross-Modal context fusion and adaptive graph convolutional neural networks for multimodal emotion recognition. The model learns cross-modal representations between pairs of three input modalities to achieve modality alignment and complementarity, enriching the input feature representation, and integrating them in the flow of contextual information. The dialogue relationship dependency graph is constructed based on the mutual and self-dependence between speakers, learning the dialogue relationship features between speakers. High detection performance was achieved on two benchmark ERC datasets.\textbf{Future work.} We will focus on designing more advanced feature fusion methods and integrating the semantic understanding capabilities of large language models to enhance the model's inference ability.

\bibliographystyle{IEEEtran}
\bibliography{reference}

\end{document}